\pdfoutput=1
\documentclass[11pt]{article}

\usepackage[]{naacl2021}

\usepackage{times}
\usepackage{latexsym}

\usepackage[T1]{fontenc}
\usepackage[utf8]{inputenc}

\usepackage{microtype}

\usepackage{booktabs,subcaption,amsmath}
\usepackage{tikz}
\usepackage{tikz-qtree}
\usepackage{enumitem} % compact item
\usepackage{fancyvrb}

\title{Non-Autoregressive Semantic Parsing for \\ 
Compositional Task-Oriented Dialog}

\author{Arun Babu \quad\quad Akshat Shrivastava 
\quad\quad Armen Aghajanyan \\ {\bf Ahmed Aly} \quad\quad {\bf Angela Fan} \quad\quad {\bf Marjan Ghazvininejad} \\
        Facebook \\
  \tt{\{arbabu, akshats, armenag, ahhegazy, angelafan, ghazvini\}@fb.com}}
\newcommand\modelname{LightConv Pointer}
\newcommand{\masktoken}{\textless MASK\textgreater}

\begin{document}
\maketitle
\begin{abstract}
Semantic parsing using sequence-to-sequence models allows parsing of deeper representations compared to  traditional word tagging based models. In spite of these advantages, widespread adoption of these models for real-time conversational use cases has been stymied by higher compute requirements and thus higher latency. In this work, we propose a non-autoregressive approach to predict semantic parse trees with an efficient seq2seq model architecture. By combining non-autoregressive prediction with convolutional neural networks, we achieve significant latency gains and parameter size reduction compared to traditional RNN models. Our novel architecture achieves up to an 81\% reduction in latency on TOP dataset and retains competitive performance to non-pretrained models on three different semantic parsing datasets. Our code is available at \url{https://github.com/facebookresearch/pytext}.
\end{abstract}

\section{Introduction}

Advances in conversational assistants have helped to improve the usability of smart speakers and consumer wearables for different tasks. Semantic parsing is one of the fundamental components of these assistants and it helps to convert the user input in natural language to a structure representation that can be understood by downstream systems. Majority of the semantic parsing systems deployed on various devices, rely on server-side inference because of the lower compute/memory available on these edge devices. This poses a few drawbacks such as flaky user experience with spotty internet connectivity and compromised user data privacy due to the dependence on a centralized server to which all user interactions are sent to. Thus, semantic parsing on-device has numerous advantages.

For the semantic parsing task, the meaning representation used decides the capabilities of the system built. Limitations of the representation with one intent and slot labels were studied in the context of nested queries and multi turn utterances in \citet{decoupled} and \citet{gupta2018semantic}. New representations were proposed to overcome these limitations and sequence-to-sequence models were proposed as the solution to model these complex forms. But using these new models in real-time conversational assistants still remains a challenge due to higher latency requirements. In our work, we propose a novel architecture and generation scheme to significantly improve the end2end latency of sequence-to-sequence models for the semantic parsing task.

Due to the autoregressive nature of generation in sequence-to-sequence semantic parsing models, the recurrence relationship between target tokens creates a limitation that decoding cannot be parallelized. 

There are multiple works in machine translation which try to solve this problem. These approaches relax the decoder token-by-token generation by allowing multiple target tokens to be generated at once. Fully non-autoregressive models \cite{gu2017non, ma2019flowseq,ghazvininejad2020aligned,saharia2020non} and conditional masked language models with iterative decoding \cite{ghazvininejad2019maskpredict,gu2019levenshtein,ghazvininejad2020semi} are some of them.

To enable non-autoregressive generation in semantic parsing, we modify the objective of the standard seq2seq model to predict the entire target structure at once. We build upon the CMLM (Conditional Masked Language Model) \cite{ghazvininejad2019maskpredict} and condition the generation of the full target structure on the encoder representation. By eliminating the recurrent relationship between individual target tokens, the decoding process can be parallelized. While this drastically improves latency, the representation of each token is still dependent on previous tokens if we continue to use an RNN architecture. Thus, we propose a novel model architecture for semantic parsing based on convolutional networks \cite{wu2019pay} to solve this issue.

% Our work focuses on reducing the latency of seq2seq semantic parsing models while maintaining state of the art accuracy on multiple benchmarks. We achieve this by augmenting non-autoregressive generation with an efficient, convolutional architecture. This allows for maximal parallelization during decoding, making it possible to use neural seq2seq models in limited compute environments such as edge-devices and substantially reduces the latency for online inference. We focus our results in a setting without pretraining as most of the large pre-trained models are not suitable for usecases under strict latency and memory budget.

Our non-autoregressive model achieves up to an 81\% reduction in latency on the TOP dataset \cite{gupta2018semantic}, while achieving 80.23\% exact match accuracy. We also achieve 88.16\% exact match accuracy on DSTC2 \cite{dstc2} and 80.86\% on SNIPS \cite{coucke2018snips} which is competitive to prior work without pretraining. 

To summarize, our two main contributions are:
\begin{itemize}
    \item  We propose a novel alternative to the traditional autoregressive generation scheme for semantic parsing using sequence-to-sequence models. With a new model training strategy and generation approach, the semantic parse structure is predicted in one step improving parallelization and thus leading to significant reduction in model latency with minimal accuracy impact. We also study the limitations of original CMLM \cite{ghazvininejad2019maskpredict} when applied for conversational semantic parsing task and provide motivations for our simple yet critical modifications.
    \item We propose \modelname{}, a model architecture for non-autoregressive semantic parsing, using convolutional neural networks which provides significant latency and model size improvements over RNN models. Our novel model architecture is particularly suitable for limited compute use-cases like on-device conversational assistants.
\end{itemize}

% To summarize, our two main contributions are: (1)  We propose a novel alternative to the traditional autoregressive generation scheme for semantic parsing using sequence-to-sequence models. With a new model training strategy and generation approach, the semantic parse structure is predicted in one step improving parallelization and thus leading to significant reduction in model latency with minimal accuracy impact. We also study the limitations of original CMLM \cite{ghazvininejad2019maskpredict} when applied for conversational semantic parsing task and provide motivations for our simple yet critical modifications. (2) We propose \modelname{}, a model architecture for non-autoregressive semantic parsing, using convolutional neural networks which provides significant latency and model size improvements over RNN models. Our novel model architecture is particularly suitable for limited compute use-cases like on-device conversational assistants.

\begin{figure}[h]
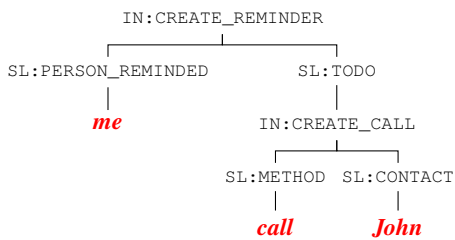

   \centering
      \centering\small
      \tikzset{level distance=20pt,sibling distance=0pt}
      \tikzset{edge from parent/.style={draw, edge from parent path={(\tikzparentnode.south) -- +(0,-4pt) -| (\tikzchildnode)}}}
      \Tree [.\texttt{\scriptsize IN:CREATE\_REMINDER} [.\texttt{\scriptsize SL:PERSON\_REMINDED} \emph{\textcolor{red}{\textbf{me}}} ] [.\texttt{\scriptsize SL:TODO} [.\texttt{\scriptsize IN:CREATE\_CALL} [.\texttt{\scriptsize SL:METHOD} \emph{\textcolor{red}{\textbf{call}}} ] [.\texttt{\scriptsize SL:CONTACT} \emph{\textcolor{red}{\textbf{John}}} ] ] ] ]
   \caption{Decoupled semantic representation for the single utterance \textit{``Please remind me to call John''.}}\label{fig:rep-single-utt}
   \label{fig:decoupled}
\end{figure}
\section{Method}

In this section, we propose a novel, convolutional, non-autoregressive architecture for semantic parsing. While non-autoregressive decoding has been previously explored in machine translation, we describe how it can be applied to semantic parsing with several critical modifications to retain performance. We then describe our convolutional architecture. By incorporating these advances together, our approach achieves both high accuracy and efficient decoding.  

The task is to predict the semantic parse tree given the raw text. We use the decoupled representation \cite{decoupled}, an extension of the compositional form  proposed in \citet{gupta2018semantic} for task oriented semantic parsing. Decoupled representation is obtained by removing all text in the compositional form that does not appear in a leaf slot. Efficient models require representations which are compact, with least number of tokens, to reduce number of floating point operations during inference. Decoupled representation was found to be suitable due to this.

Figure~\ref{fig:rep-single-utt} shows the semantic parse for a sample utterance. Our model predicts the serialized representation of this tree which is 
% $[IN:CREATE\_REMINDER\ [SL:PERSON\_REMINDED\ me\ ]\ [SL:TODO\ [IN:CREATE\_CALL\ [SL:METHOD\ call\ ]\ [SL:CONTACT\ John\ ]\ ]\ ]\ ]$
\begin{Verbatim}[fontsize=\small]
  [IN:CREATE_REMINDER [SL:PERSON_REMINDED me ]
  [SL:TODO [IN:CREATE_CALL [SL:METHOD call ]
  [SL:CONTACT John ] ] ] ]
\end{Verbatim}

\subsection{Non-Autoregressive Decoding}
\label{sec:nar_decoding_motivations}
\begin{figure}[t]
\includegraphics[width=\columnwidth]{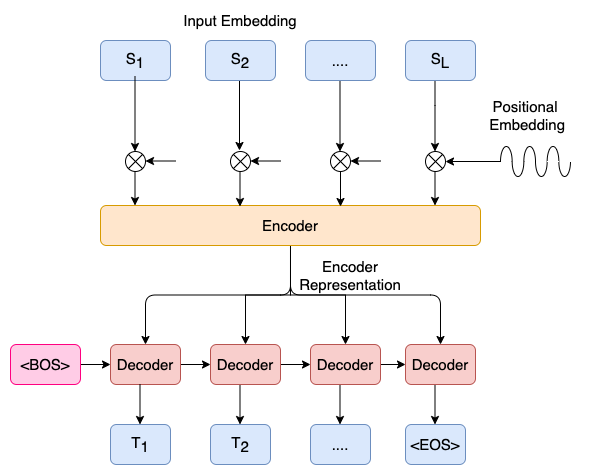}
\caption{Traditional Sequence to Sequence architecture which uses autoregressive generation scheme for decoder. } 
\label{fig:ar_block}
\end{figure}

While autoregressive models (Figure \ref{fig:ar_block}), which predict a sequence token by token, have achieved strong results in various tasks including semantic parsing, they have a large downside. The main challenge in practical applications is the slow decoding time. We investigate how to incorporate recent advances in non-autoregressive decoding for efficient semantic parsing models. 

We build upon the Conditional Masked Language Model (CMLM) proposed in \citet{ghazvininejad2019maskpredict} by applying it to the structured prediction task of semantic parsing for task-oriented dialog. \citet{ghazvininejad2019maskpredict} uses CMLM to first predict a token-level representation for each source token and a target sequence length; then the model predicts and iterates on the target sequence prediction in a non-autoregressive fashion. We describe our changes and the motivations for these changes below.

One of the main differences between our work and \citet{ghazvininejad2019maskpredict} is that target length prediction plays a more important role in semantic parsing. For the translation task, if the target length is off by one or more, the model can slightly rephrase the sentence to still return a high quality translation. In our case, if the length prediction is off by even one, it will lead to an incorrect semantic parse.

To resolve this important challenge, we propose a specialized length prediction module that more accurately predicts the target sequence length. While \citet{ghazvininejad2019maskpredict} uses a special CLS token in the source sequence to predict the target length, we have a separate module of multiple layers of CNNs with gated linear units to predict the target sequence length \citep{wu2019pay}. We also use label smoothing and differently weighing losses as explained in section  \ref{sec:modeltraining}, to avoid the easy overfitting in semantic parsing compared to translation.

As shown in \citet{decoupled}, transformers without pre-training perform poorly on TOP dataset. The architectural changes that we propose to solve the data efficiency can be found in the section \ref{sec:model_architecture}.

Further, we find that the random masking strategy proposed in \citet{ghazvininejad2019maskpredict} works poorly for semantic parsing. When we use the same strategy for the semantic parsing task where the output has a structure, model is highly likely to see invalid trees during training as masking random tokens in the linearized representation of a tree mostly gives invalid tree representations. This makes it hard for the model to learn the structure especially when the structure is complicated (in the case of trees, deep trees were harder to learn). To remedy this problem, we propose a different strategy for model training where all the tokens in the target sequence are masked during training.

Ablation experiments for all the above changes can be found in section \ref{sec:ablation_explanation}.

\subsection{\modelname{} Model}
\subsubsection{Model Architecture}
\label{sec:model_architecture}

\begin{figure}[t]
\centering
\includegraphics[width=0.90\columnwidth]{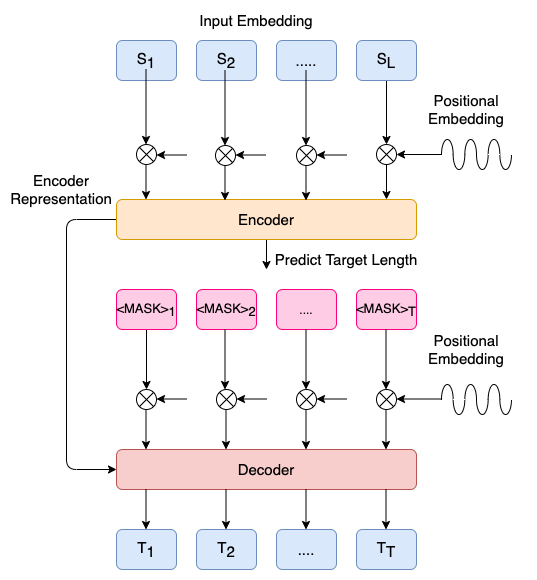}
\caption{Sequence to Sequence model architecture which uses Non-Autoregressive strategy for generation}
\label{fig:nar_block}
\end{figure}

Our model architecture (Figure \ref{fig:nar_block}) is based on the classical seq2seq model \cite{sutskever2014sequence} and follows the encoder-decoder architecture. In order to optimize for efficient encoding and decoding, we look to leverage a fully parallel model architecture. While transformer models are fully parallel and popular in machine translation \cite{Transformer}, they are known to perform poorly in low resource settings and require careful tuning using techniques like Neural Architecture Search to get good performance \cite{biljon2020optimal,murray-etal-2019-auto}. Similarly, randomly initialized transformers performed poorly on TOP dataset achieving only 64.5 \% accuracy when SOTA was above 80\% \cite{decoupled}. We overcome this limitation by augmenting Transformers with Convolutional Neural Networks. Details of our architecture are explained below.
 
\begin{figure}[t]
\centering
\includegraphics[width=0.9\columnwidth]{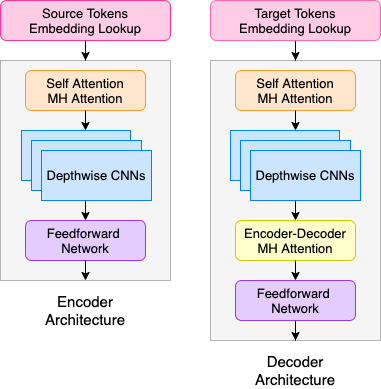}
\caption{Different layers in \modelname{} Model Architecture}
\label{fig:arch_layers}
\end{figure}

For token representations, we use word embeddings concatenated with the sinusoidal positional embeddings \cite{Transformer}. Encoder and decoder consist of multiple layers with residual connections as shown in Figure \ref{fig:arch_layers}.

First sub-block in each layer consists of MHA  \cite{Transformer}. In decoder, we do not do masking of future tokens during model training. This is needed for non-autoregressive generation of target tokens during inference.

Second sub-block consists of multiple convolutional layers. We use depthwise convolutions with weight sharing \cite{wu2019pay}. Convolution layer helps in learning representation for tokens for a fixed context size and multiple layers helps with bigger receptive fields. We use non-causal convolutions for both encoder as well as decoder.

Third sub-block is the FFN \cite{Transformer,wu2019pay} which consists of two linear layers and relu. The decoder has source-target attention after the convolution layer.

\textbf{Pointer-Generator Projection layer}
The decoder has a final projection layer which generates the target tokens from the decoder/encoder representations. \citet{rongali2020don} proposes an idea based Pointer Generator Network \cite{see2017get} to convert the decoder representation to target tokens using the encoder output. Similarly, we use a pointer based projection head, which decides whether to copy tokens from the source-sequence or generate from the pre-defined ontology at every decoding step \cite{decoupled}.

\textbf{Length Prediction Module}
Length prediction Module receives token level representations from the encoder as input. It uses stacked CNNs with gated linear units and mean pooling to generation the length prediction.

\subsubsection{Inference}

Suppose the source sequence is of length L and source tokens in the raw text are $s_1, s_2, s_3 \dotsc s_L$. Encoder generates a representation of for each token in the source sequence.

\begin{equation}
    e_1, \dotsc, e_L = \operatorname{Encoder}(s_1, \dotsc, s_L)
\end{equation}

The length prediction module predicts the target sequence length using the token level encoder representation.

\begin{equation}
    T = \operatorname{PredictLength}(e_1, \dotsc, e_L)
\end{equation}

Using the predicted length T, we create a target sequence of length T consisting of identical MASK tokens. This sequence is passed through possibly multiple decoder layers and generates a representation for each token in the masked target sequence.

\begin{equation}
    x_1, ..., x_T =  \operatorname{Dec}(\text{MASK}_1,…, \text{MASK}_T; e_1, ..., e_L)
\end{equation}

We use \textbf{Pointer-Generator Projection layer} explained in \ref{sec:model_architecture} to predict target tokens.

\begin{equation}
    y_1, ..., y_T = \operatorname{PtrProj}(x_1, .., x_T; e_1, .., e_L)
\end{equation}

We make a strong assumption that each token in the target sentence is conditionally independent of each other given the source and the target length. Thus, the individual probabilities for each token is $P(y_i | X, T)$ where $X$ is the input sequence and $T$ is the length of target sequence.

\paragraph{Beam Search} During inference, length prediction module explained in  \ref{sec:model_architecture} predicts top \textit{k} lengths. For each predicted length, we create a decoder input sequence of all masked tokens. This is similar to the beam search  with beam size \textit{k} in autoregressive systems. The main difference in our model architecture is that we expect only one candidate for each predicted length. These all masked sequences are given as input to the model and the model predicts target tokens for each masked token. Once we have predicted target sequences for \textit{k} different lengths, they are ranked based on the ranking algorithm described in (\ref{eq:ranking}), where $X$ is the input sequence and $Y$ is the predicted output sequence, note the predicted token $y_i$ is conditioned on both the sequence ($X$) and the predicted target length $T$.

\begin{equation}
    \label{eq:ranking}
    S(X, Y) = \sum_{i}^{T}P(y_i \mid X, T) \cdot P(T)
\end{equation}

\subsection{Training}

\label{sec:modeltraining}

During training, we jointly optimize for two weighted losses. The first loss is calculated for the predicted target tokens against the real target and the second loss is calculated for predicted target length against real target length.

During forward-pass, we replace all the tokens in the target sequence with a special \masktoken{}  token and give this as an input to the decoder. Decoder predicts the token for each masked token and the cross-entropy loss is calculated for each predicted token.

The length prediction module in the model predicts the target length using the encoder representation. Similar to CMLMs in \cite{ghazvininejad2019maskpredict}, length prediction is modeled as a classification task with class labels for each possible length. Cross entropy loss is calculated for length prediction. For our semantic parsing task, label smoothing \cite{szegedy2015rethinking} was found to be very critical as the length prediction module tends to easily overfit and strong regularization methods are needed. This was because length prediction was a much well-defined task compared to predicting all the tokens in the sequence.

Total loss was calculated by taking a weighted sum of cross entropy loss for labels and length, with lower weight for length loss.

As training progresses through different epochs, the best model is picked by comparing the exact match (EM) accuracy of different snapshots on validation set. 
% Loss changes were found to be quite noisy and occasionally ended in premature termination of model training when used as the condition to pick best snapshots.

% \textbf{Knowledge distillation}
% \label{sec:training_distillation}
% Following prior work \cite{ghazvininejad2019maskpredict, Zhou2020Understanding}, we train our model with sequence-level knowledge distillation~\cite{kim2016sequence}. We train our system on data generated by the current SOTA autoregressive models BART~\cite{lewis2019bart}. In Table~\ref{table:results}, we show the importance of knowledge distillation in our task and also report the result of an autoregressive model on the same distilled data as reference.

\section{Experiments}
\subsection{Datasets}
\label{sec:datasets}
We use 3 datasets across various domains to evaluate our semantic parsing approach. Length distribution of each dataset is described using median, 90th percentile and 99th percentile lengths.

\textbf{TOP Dataset} Task Oriented Parsing \cite{gupta2018semantic} is a dataset for compositional utterances in the navigation and events domains. The training set consists of $31,279$ instances and the test set consists of  $9,042$. The test set has a  median target length of 15, P90 27 and P99 39.

\textbf{SNIPS} The SNIPS \cite{coucke2018snips} dataset is a public dataset used for benchmarking semantic parsing intent slot models. This dataset is considered flat, since it does not contain compositional queries and can be solved with word-tagging models. Recently, however seq2seq models have started to out perform word-tagging models \cite{rongali2020don, decoupled}. The training set consists of $13,084$ instances, the test set consists of $700$ instances. The test set has a median target length of 11, P90 17, P99 21.

\textbf{DSTC2} Dialogue State Tracking Challenge 2 \cite{dstc2}, is a dataset for conversational understanding. The dataset involves users searching for restaurants, by specifying constraints such as cuisine type and price range, we encode these constraints as slots and use this to formulate the decoupled representation. The training set consists of $12,611$ instances  and a test set of $9890$. The test set has a median target length of 6, P90 9 and P99 10.

\begin{figure*}
        \centering
        \subfloat[Exact Match Accuracy on TOP, DSTC2, and SNIPS]{
            \small
            \begin{tabular}{@{}lrrr@{}}
\toprule
                        & \multicolumn{3}{c}{Exact Match Accuracy}                         \\ \midrule
Model                   & \textbf{TOP}  & \textbf{DSTC2} & \textbf{SNIPS} \\ \midrule
RNNG  \cite{einolghozati2018improving}     & 80.86                      &     -           &   -            \\ \midrule

Ptr Transformer \cite{rongali2020don}    &  79.25                       &    -            &    85.43           \\ \midrule
Ptr BiLSTM  \cite{decoupled}     &  79.51                       &    88.33            &    -           \\ \midrule
GLAD \cite{glad}     & -                      &     79.4           &   -            \\ \midrule
JointBiRNN \cite{hakkani2016multi}     &  -                       &    -            &    73.20           \\ \midrule
Slot Gated \cite{goo2018slot}     &  -                       &    -            &    75.50           \\ \midrule
Capsule NLU \cite{zhang2018joint}     &  -                       &    -            &    80.90           \\ \midrule
\toprule  &  \multicolumn{3}{c}{Ours} \\ \midrule  
NAR \modelname{}        & 80.20                 & 88.16       & 80.86        \\  \midrule
AR \modelname{}          & 80.23              &   88.58             & 76.43              \\ \midrule
\bottomrule
\end{tabular}
\label{table:results}
            }
        \subfloat[Median latency on TOP dataset]{           
            \includegraphics[width=7.5cm]{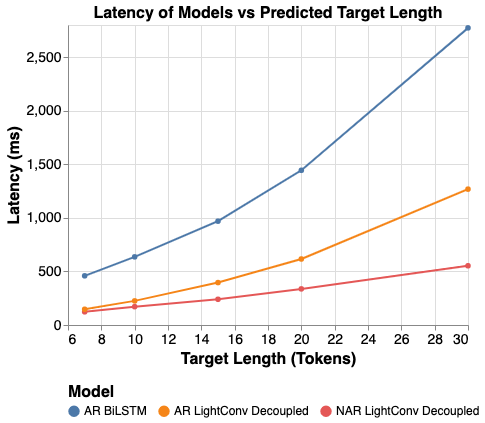}

            \label{fig:latency}      

            }
    \caption{\small \textbf{(a)}: Exact match (EM) accuracy is shown on the test set across 3 different datasets. We compare our proposed \modelname{} variants (AR and NAR) against various baselines that do not include pre-trained representations. \textbf{(b)}: Median latency of the NAR \modelname{}, AR \modelname{}, and the Seq2Seq Pointer BiLSTM baseline (termed Ptr BiLSTM in figure \ref{table:results}) \cite{decoupled} varying over increasing target sequence length on the TOP dataset.}
    \end{figure*}

\subsection{Evaluation}
\begin{table*}[h]
\centering
\small
\begin{tabular}{@{}lllrrrrrr@{}}
\toprule
& & & & \multicolumn{5}{c}{Mean Length Bucket Exact Match Accuracy (\%)} \\  
Model & & & EM  & \textless{} 10 EM    &  10-20 EM  &  20-30 EM  & \textgreater{} 30 EM                 \\ \hline \\
CMLM Transformer & + CLS & + Random Masking & 70.9        & 79.3  & 74.5 & 35.1 & 0.4     \\
CMLM LightConv & + CLS & + Random Masking      & 78.3 & 82.9 & 81.8 & 53.5 & 3.9                              \\ 
CMLM LightConv & + Conv Length & + Random Masking         & 79.4 & 83.3 & 82.5 & 58.2 & 5.1       \\ 
CMLM LightConv & + CLS & + Mask Everything         & 78.6 & 82.7 & 81.8 & 56.0 & 9.4       \\ 
CMLM LightConv & + Conv Length & + Mask Everything         & 79.6 & 82.2 & 82.8 & 61.4 & 14.9       \\ 
\bottomrule
\end{tabular}
\caption{Ablation experiments reporting EM in different buckets based on the target sequence length. Bucket sizes are 2798, 5167, 992 and 85 respectively. It can be seen the our model setup works significantly better, especially for longer sequences.}
\label{table:ablation}
\end{table*}

\label{sec:evaluation}
\paragraph{Semantic Parsing Performance}
For all our datasets, we convert the representation of either the compositional form or flat intent slot form to the decoupled representation \cite{decoupled} . We compare the model prediction with the serialized structure representation and look for exact match (EM).

\paragraph{Benchmarking Latency}
For the latency analysis for the models trained from scratch: AR \modelname{}, NAR \modelname{}, and BiLSTM. We chose these 3 architectures, to compare NAR vs AR variants of \modelname{}, as well as the best performant baseline: Pointer BiLSTM \cite{decoupled}. We use Samsung Galaxy S8 with Android OS and Octa-core processor. We chose to benchmark latency to be consistent with prior work on on-device modeling \cite{wu2019fbnet, howard2019searching}. All models are trained in PyTorch \cite{NEURIPS2019_PyTorch} and exported using Torchscript. We measure wall clock time as it is preferred instead of other options because it relates more to real world inference. \footnote{We use the open source framework https://github.com/facebook/FAI-PEP for latency benchmarking.} Latency results can be found in section \ref{sec:latency_analysis}.

\subsection{Baselines}
\label{sec:baselines}
For each of our datasets, we report accuracy metrics on the following models:
\begin{itemize}
    \item \textbf{AR \modelname{}}: Autoregressive (AR) \modelname{} model to establish an autoregressive baseline of our proposed architecture.
    \item \textbf{NAR \modelname{}}: A non-autoregressive (NAR) variant of the above model to allow for parallel decoding.
\end{itemize}

We compare against the best reported numbers across datasets where the models don't use pre-training.

\subsection{Model Training Details}
\label{sec:model_training}
During training of our model we use the same base model across all datasets and sweep over hyper parameters for the length module and the batch size and learning rate, an equivalent sweep was done for the AR variant as well. The base model we use for NAR \modelname{} model uses 5 encoder layers  with convolutional kernel sizes [3,7,15,21,27], where each encoder layer has embedding and convolutional dimensions of 160, 1 self attenion head, and 2 decoder layers with kernel sizes [7,27], and embedding dimension of 160, 1 self-attention head and 2 encoder-attention heads. Our length prediction module leverages a two convolution layers of 512 embedding dimensions and kernel sizes of 3 and 9. and uses hidden dimension in [128,256,512] determined by hyper parameter sweeps. We also use 8 attention heads for the decoupled projection head. For the convolutional layer, we use lightweight convolutions \cite{wu2019pay} with number of heads set to 2. We train with the Adam \cite{kingma2014adam} optimizer, learning rate is selected to be between [0.00007, 0.0004]. If our evaluation accuracy has not increased in 10 epochs, we also reduce our learning rate by a factor of 10, and we employ early stopping if the accuracy has not changed in 20 epochs. We train with our batch size fixed to be 8. We optimize a joint loss for label prediction and length prediction. Both losses consist of label smoothed cross entropy loss ($\beta$ is the weight of the uniform distribution) \cite{pereyra2017regularizing}, our label loss has $\beta = 0.1$ and our length loss has $\beta = 0.5$, we also weight our length loss lower, $\lambda = 0.25$. For inference, we use a length beam size of $k=5$. Our AR variant follows the same parameters however it does not have length prediction and self-attention in encoder and decoder.

\section{Results}
\label{sec:results}
We show that our proposed non-autoregressive convolutional architecture for semantic parsing is competitive with auto-regressive baselines and word tagging baselines without pre-training on three different benchmarks and reduces latency up to 81\% on the TOP dataset. We first compare accuracy and latency, then discuss model performance by analyzing errors by length, and the importance of knowledge distillation. We do our analysis on the TOP dataset, due to its inherent compositional nature, however we expect our analysis to hold for other datasets as well. Non-compositional datasets like DSTC2 and SNIPS can be modeled by word tagging models making seq2seq models more relevant in the case of compositional datasets.

\subsection{Accuracy}

% In table \ref{table:results} we show our NAR and AR variants for \modelname{} perform quite similarly across all datasets. While NAR lags slightly behind AR approaches, knowledge distillation from AR BART further improves performance across all datasets. We can see that our proposed NAR \modelname{} is also competitive with state of the art models without pre-training before knowledge distillation: -0.66\% TOP, -0.17\% DSTC2, -4.57\% SNIPS  (-0.04\% compared to word tagging models), and further improves some datasets after distillation (+0.69\% TOP, +0.0\% on DSTC2, +0.85\% SNIPS compared to the non distilled baseline). However, we do note that the best reported models do not leverage knowledge distillation. 
In table \ref{table:results} we show our NAR and AR variants for \modelname{} perform quite similarly across all datasets. We can see that our proposed NAR \modelname{} is also competitive with state of the art models without pre-training: -0.66\% TOP, -0.17\% DSTC2, -4.57\% SNIPS  (-0.04\% compared to word tagging models). Following the prior work on Non-Autoregressive models, we also report our experiments with sequence-level knowledge distillation in subsection Knowledge Distillation under section. \ref{sec:ablation_explanation}.
\subsection{Latency}
\label{sec:latency_analysis}

In figure \ref{fig:latency} we show the latency of our model with different generation approaches (NAR vs AR) over increasing target sequence lengths on the TOP dataset. Firstly, we show that our \modelname{} is significantly faster than the BiLSTM baseline \cite{decoupled}, achieving up to a 54\% reduction in median latency. BiLSTM was used as baseline as that was the SOTA without pretraining for TOP and Transformers performed poorly. By comparing our model with AR and NAR generation strategy, it can be seen that increase in latency with increase in target length is much smaller for NAR due to better parallelization of decoder, resulting in up to an 81\% reduction in median latency compared to the BiLSTM model. Also note that both the \modelname{} models are able to achieve parity in terms of EM Accuracy compared to the baseline BiLSTM model, while using many fewer parameters, the BiLSTM model uses 20M parameters, while the NAR \modelname{} uses 12M and the AR \modelname{} uses 10M.

\subsection{Analysis}
\label{sec:ablation_explanation}
\paragraph{Ablation experiments}
We compare the modifications proposed by this work (LightConv, Conv length prediction module and Mask everything strategy) with the original model proposed in \citet{ghazvininejad2019maskpredict} in table \ref{table:ablation}. The motivations for each modification was already discussed in sub-section \ref{sec:nar_decoding_motivations}. Our mean EM accuracy results based on 3 trials show the significance of techniques proposed in this paper especially for longer target sequences.

\paragraph{Errors by length}

\begin{table}[]
\centering
\small
\begin{tabular}{@{}lrrrr@{}}
\toprule
Length Bucket & NAR (\%) & AR (\%) & Bucket Size                 \\ \hline \\
\textless{} 10 & 82.80        & 83.13       & 2798 \\
10-20         & 84.18        & 84.36       & 5167                        \\ 
20-30         & 62.50        & 65.72      & 992  \\ 
30-40         & 21.25        & 41.25      & 80                          \\
\textgreater{} 40           & 0.00         & 20.00       & 5                           \\ \bottomrule
\end{tabular}
\caption{EM accuracy of the NAR \modelname{} (distilled) vs AR \modelname{} distilled across different target length buckets along with the number of instances in each bucket on the TOP dataset.}
\label{table:length_bucket}
\end{table}

It is known that non-autoregressive models have difficulty at larger sequence lengths \cite{ghazvininejad2019maskpredict}. In table \ref{table:length_bucket}, we show our model's accuracy in each respective length bucket on the TOP dataset. We see that the AR and NAR model follow a similar distribution of errors, however the NAR model seems to error at a higher rate for the longer lengths. 
% This is consistent with our expectation that totally non-autoregressive generation is much harder as sequence length grows. 
% We also note that since this is the tail end of the distribution, it may require more data points to train the non-autoregressive variant at these longer lengths. We tried oversampling approaches to encourage the model to learn more at these longer lengths, but got no significant gains.

\paragraph{Knowledge Distillation}
Following prior work \cite{ghazvininejad2019maskpredict, Zhou2020Understanding}, we train our model with sequence-level knowledge distillation~\cite{kim2016sequence}. We train our system on data generated by the current SOTA autoregressive models BART~\cite{lewis2019bart, decoupled}. In table \ref{table:kd_res} we show the impact of knowledge distillation in our task on both the non-autoregressive and autoregressive variants of \modelname{}. These results support prior work in machine translation for distillation of autoregressive teachers to non-autoregressive models showing distillation improving our models on TOP and SNIPS, however we notice minimal changes on DSTC2.

\paragraph{The importance of length prediction}

\begin{figure}[t]
\centering
\includegraphics[width=6cm]{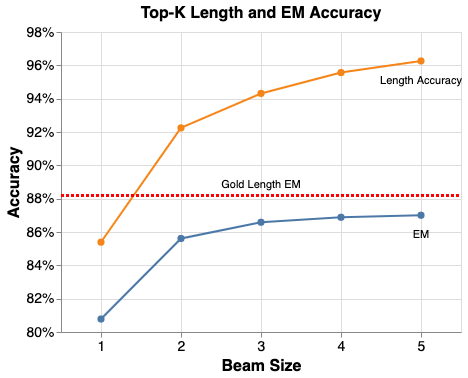}
\caption{Distilled NAR \modelname{} Top-K accuracy for exact match (EM) accuracy (blue) and Top-K length accuracy (orange), as well as the EM accuracy with gold length (dotted red line) for the TOP dataset.}
\label{fig:beam_predictions}
\vspace{-5pt}
\end{figure}

An important part of our non-autoregressive model is length prediction. In figure \ref{fig:beam_predictions}, we report exact match accuracy @ top k beams and length accuracy @ top k beams (where top K refers to whether the correct answer was in the top K predictions) for the TOP dataset. We can see a tight correlation between our length accuracy and exact match accuracy, showing how our model is bottle necked by the length prediction.

% Looking at our results, we hypothesized that we should be able to match our top-5 accuracy, if we provided gold length as a feature. 
Providing gold length as a feature, led to an exact match accuracy of \textbf{88.20\%} (shown in red on figure \ref{fig:beam_predictions}), an absolute 7.31 point improvement over our best result with our non-autoregressive \modelname{}.  

\begin{table}[]
\centering
\small
\begin{tabular}{@{}lrrr@{}}
\toprule
Model & TOP & DSTC2  & SNIPS         \\ \hline \\
BART (Teacher Model) & 87.10        & 89.06 &    91.00    \\
Distilled NAR \modelname{}      & 80.89        & 88.16 & 81.71                              \\ 
Distilled AR \modelname{}         & 81.53       & 88.21 & 80.29       \\ 
\bottomrule
\end{tabular}
\caption{EM accuracy of various models leveraging KD from the teacher BART model on the TOP dataset.}
\label{table:kd_res}
\vspace{-5pt}
\end{table}

\section{Related Work}
\textbf{Non-autoregressive Decoding}
Recent work in machine translation has made a lot of progress in fully non-autoregressive models \cite{gu2017non, ma2019flowseq,ghazvininejad2020aligned,saharia2020non}  and parallel decoding \cite{lee2018,ghazvininejad2019maskpredict, gu2019levenshtein,ghazvininejad2020semi,kasai2020parallel}.  
While many advancements have been made in machine translation, we believe we are the first to explore the non-autoregressive semantic parsing setting. In our work, we extend the CMLM to work for semantic parsing. We make two important adjustments: first, we use a different masking approach where we mask everything and do one-step generation. Second, we note the importance of the length prediction task for parsing and improve the length prediction module in the CMLM.

% \textbf{Efficient Modeling}
% Efficient modeling has peaked interests of researchers as they look for privacy preserving AI and on device models. 
% Previous work has focused on designing efficient architectures \cite{tay2019lightweight,cao2017mobirnn,li2016lightrnn}, however they were not focused on semantic parsing. We take inspiration from \citealt{desai2020lightweight} and \citealt{wu2019pay} to use CNNs instead of RNN based models as it results in faster and smaller models while still being performant.

\textbf{Seq2Seq For Semantic Parsing}
Recent advances in language understanding have lead to increased reliance on seq2seq architectures. Recent work by \citealt{rongali2020don, decoupled}, showed the advantages from using a pointer generator architecture for resolving complex queries (e.g. composition and cross domain queries) that could not be handled by word tagging models. Since we target the same task, we adapt their pointer decoder into our proposed architecture. However, to optimize for latency and compression we train CNN based architectures (\citealt{desai2020lightweight} and \citealt{wu2019pay}) to leverage the inherent model parallelism compared to the BiLSTM model proposed in \citealt{decoupled} and more compression compared to the transformer seq2seq baseline proposed in \citealt{rongali2020don}. To further improve latency we look at parallel decoding through non-autoregressive decoding compared to prior work leveraging autoregressive models.

\section{Conclusion}
This work introduces a novel alternative to autoregressive decoding and efficient encoder-decoder architecture for semantic parsing. We show that in 3 semantic parsing datasets, we are able to speed up decoding significantly while minimizing accuracy regression. Our model is able to generate parse trees competitive with state of the art auto-regressive models with significant latency savings, allowing complex NLU systems to be delivered on edge devices. 

There are a couple of limitations of our proposed model that naturally extend themselves to future work. Primarily, we cannot support true beam decoding, we decode a single prediction for each length prediction however there may exist multiple beams for each length prediction. Also for longer parse trees and more complex semantic parsing systems such as session based understanding, our NAR decoding scheme could benefit from multiple iterations. Lastly, though we explored models without pre-training in this work, recent developments show the power of leveraging pre-trained models such as RoBERTa and BART. We leave it to future work to extend our non-autoregressive decoding for pre-trained models.

\section*{Acknowledgements}
We would like to thank - Sandeep Subramanian (MILA),  Karthik Prasad (Facebook AI), Arash Einolghozati (Facebook) and Yinhan Liu for the helpful discussions.

\bibliography{custom}
\bibliographystyle{acl_natbib}

\end{document}